# Loss-Guided Adaptive Scale Refinement for Molecular Force Prediction

Limin Yu[1]

[1]School of Medical Imaging, Division of Medical Technology, Tianjin Key Laboratory of Functional Imaging, Tianjin Medical University, Tianjin 300203, China

## Abstract

Molecular systems exhibit interactions across multiple spatial scales, ranging from local coordination and short-range perturbations to long-range electrostatic and solvent-mediated effects. Most molecular representation learning methods rely on manually predefined scales, such as atom-level, fragment-level, residue-level, or global representations. However, the task-optimal modeling scale may not coincide with these fixed human-defined levels. Here, this study introduces a loss-guided adaptive scale refinement framework for molecular force prediction. The central idea is to treat predefined scales as initial anchors and to discover task-effective modeling resolutions through scale interpolation, scale routing, differentiable scale updates, and scale pool refinement.

Using a NaCl aqueous ionic system as a minimal testbed, this study constructs short-scale and long-range force prediction branches and analyzes their complementarity. Oracle hard routing, which selects the better branch for each atom, reduces the overall force MAE from 399.65 to 382.67, corresponding to a 4.25% improvement over the long-range baseline. In close-contact regimes with nearest-ion distance below 0.6 nm, the improvement reaches 17.35%. More importantly, continuous oracle interpolation between the short- and long-range branches further reduces the overall MAE to 380.96 and the close-contact MAE to 260.51, indicating that the optimal scale is not always located at manually defined short/long endpoints. A minimal scale pool update experiment further shows that starting from endpoint anchors {0,1}, loss-guided updates automatically generate intermediate scale anchors and recover most of the continuous oracle performance. The final updated scale pool {0,0.125,0.25,0.375,0.5,0.75,1} achieves an overall MAE of 381.23, close to the continuous oracle MAE of 380.96.

This study further evaluates learnable scale routing and differentiable scale update models. A gradient-updated MLP scale gate achieves the best non-oracle performance, reducing overall MAE to 396.55 and close-contact MAE to 298.48, corresponding to improvements of 0.78% and 8.78%, respectively. These results support adaptive scale refinement as a promising direction for molecular representation learning, especially in regimes where fixed-scale modeling is insufficient.

**Keywords:** Adaptive scale refinement; Molecular force prediction; Neural network potentials; Multiscale molecular modeling; Loss-guided scale selection; Differentiable scale update; Task-conditioned molecular modeling.

# 1. Introduction

Molecular and condensed-phase systems are intrinsically multiscale. Local interactions such as steric repulsion, coordination, hydrogen bonding, and short-range electrostatic perturbations coexist with longer-range effects including solvent organization, ionic screening, collective polarization, and global molecular context. Accurately representing these interactions is a central challenge in molecular machine learning and neural network force-field modeling. Early high-dimensional neural network potentials and transferable atomistic neural potentials demonstrated that atom-centered representations can be used to approximate potential-energy surfaces with substantially lower cost than direct electronic-structure calculations [1,2]. Subsequent molecular deep learning models, such as continuous-filter convolutional networks and physically informed neural architectures, further improved the modeling of molecular energies, forces, charges, long-range interactions, and nonlocal effects [3-5]. Graph neural networks have also been developed as a general representation framework for molecules and crystals, further supporting the use of graph-based structural representations in atomistic modeling [6,7]. More recently, geometric, directional, and equivariant graph neural networks have provided powerful representations for atomistic systems by explicitly incorporating spatial symmetry, angular information, and directional message passing [8–12].

Despite these advances, most molecular representation learning methods still rely on manually predefined modeling resolutions, such as atom-level graphs, fragment-level graphs, residue-level graphs, local cutoffs, or global molecular descriptors. These predefined scales are often effective, but they may not coincide with the task-optimal modeling scale. In practice, the most useful representation scale can vary with the target property, local chemical environment, and physical constraints. Broad molecular learning benchmarks and learned molecular representation studies have shown that prediction performance depends strongly on task type, dataset composition, and evaluation setting [13,14]. For example, force prediction near an ion may require fine local information from the first coordination shell, whereas force prediction for solvent atoms may rely more heavily on broader environmental context. Similarly, infrared spectral prediction may depend on bond- or functional-group-level vibrational patterns, docking may depend on local protein-ligand contact regions and binding-pocket environments, and toxicity prediction may depend on specific toxicophore motifs as well as whole-molecule physicochemical properties.

Therefore, the central question is not simply which fixed scale is best. Rather, given a task objective, local molecular environment, and possible physical or computational constraints, the key problem is to determine which effective modeling scale minimizes prediction error. This perspective differs from ordinary multiscale fusion. Instead of combining a set of fixed scales with predefined roles, this study treats manually defined scales as initial anchors and uses task loss to discover, interpolate, update, or generate task-effective scales.

The work studies this idea in a minimal force-prediction setting. This study uses a NaCl aqueous ionic system and constructs two initial scale experts: a short-scale branch and a long-range branch. The short-scale branch is intended to capture local coordination and close-contact effects, whereas the long-range branch is intended to capture broader electrostatic and solvent-mediated context. Rather than assuming that either branch is universally optimal, this study asks whether better predictions can be obtained through adaptive interpolation, routing, differentiable scale updates, and loss-guided scale-pool refinement.

Our main hypothesis is that manually defined molecular scales should be treated as initial anchors, while task-effective scales should be discovered or updated according to prediction loss, local environment, and complexity constraints. This hypothesis is related to conditional computation and mixture-of-experts ideas, where different inputs can be routed to different computational experts [8]. However, in the present setting, the experts are interpreted as different molecular modeling scales, and the routing or interpolation is guided by force-prediction loss and local physical environment.

This paper makes the following contributions:

1. This study formulates molecular scale selection as a loss-guided adaptive scale refinement problem.
2. This study shows that short- and long-range force prediction branches exhibit substantial complementarity, especially in close-contact ionic regimes.
3. This study demonstrates that continuous scale interpolation provides a higher oracle upper bound than hard short/long routing, supporting the existence of intermediate effective scales.
4. This study shows that a compact set of discrete scale anchors nearly approximates the continuous oracle scale space.
5. This study proposes and validates a scale pool update mechanism that automatically generates intermediate effective scales from endpoint anchors.
6. This study introduces differentiable scale update models in which scale weights are directly optimized through force prediction loss.
7. This study provides initial evidence that learnable scale gates can partially translate adaptive scale discovery into improved force prediction.
8. This study further outlines a task-conditioned scale-pool extension, in which different molecular tasks can induce different effective scale distributions.

## 2. Method

The study first provides an overview of the proposed framework in Figure 1. The framework starts from molecular inputs and two endpoint scale experts, and then performs adaptive scale refinement through interpolation, routing, differentiable scale updates, and loss-guided scale pool refinement. A task-conditioned extension is further included to generalize the framework beyond single-task force prediction.

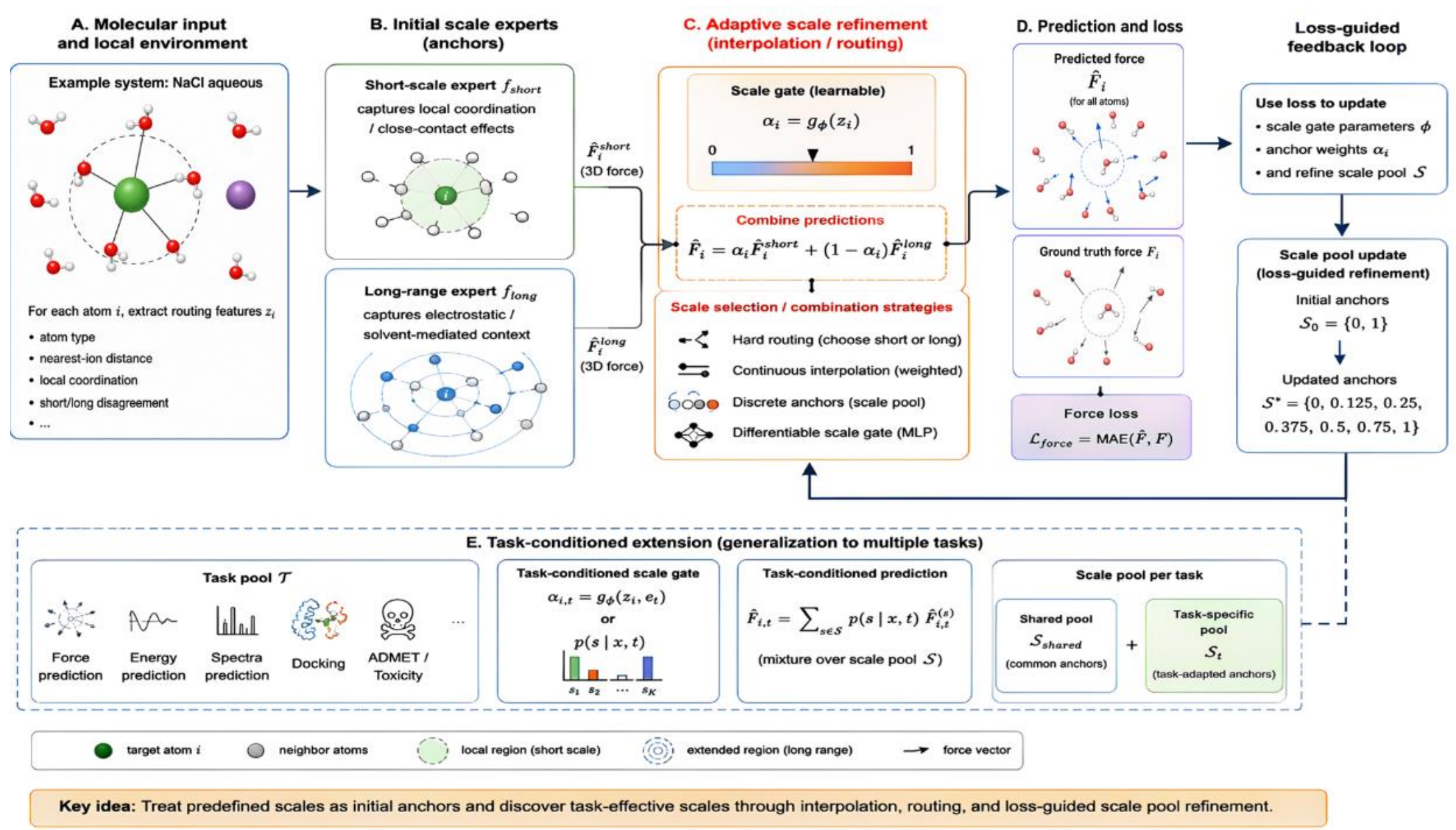

**Figure 1. Overview of the loss-guided adaptive scale refinement framework. Molecular inputs are processed by short-scale and long-range experts, whose predictions are combined through adaptive scale refinement modules, including interpolation, routing, differentiable scale updates, and loss-guided scale pool refinement. A task-conditioned extension is included to illustrate how the framework can be generalized to multiple molecular prediction tasks.**

## 2.1 Task-constrained molecular learning

This study considers a molecular prediction task defined by a supervised task loss and optional physical or chemical constraints. For a task t, the total objective can be written as

$$\mathcal{L}_t^{total} = \mathcal{L}_t^{task} + \lambda_g \sum_i g_i(x,\theta)^2 + \lambda_h \sum_j max(0, h_j(x,\theta))^2 \tag{2.1}$$

where $\mathcal{L}_t^{task}$ is the prediction loss, $g_i(x,\theta) = 0$ represents equality constraints, and $h_j(x,\theta) \leq 0$ represents inequality constraints.

This study focuses on atom-level force prediction. For atom i, the target force is $F_i = (F_{ix}, F_{iy}, F_{iz})$ and the predicted force is $\hat{F}_i = (\hat{F}_{ix}, \hat{F}_{iy}, \hat{F}_{iz})$.

The force prediction loss is defined as component-wise MAE:

$$\mathcal{L}_{force} = \frac{1}{3N}\sum_{i=1}^{N}\left(|\hat{F}_{ix} - F_{ix}| + |\hat{F}_{iy} - F_{iy}| + |\hat{F}_{iz} - F_{iz}|\right) \tag{2.2}$$

Although this study focuses on force MAE, the framework can be extended to energy-force consistency, charge conservation, stability constraints, or other task-specific physical regularizers.

## 2.2 Short- and long-range scale experts

This study defines two initial scale anchors: $s_{short}, s_{long}$.

Each scale is associated with a prediction branch: $f_{short}(x_i) \rightarrow \hat{F}_i^{short}, f_{long}(x_i) \rightarrow \hat{F}_i^{long}$.

The short-scale branch is intended to capture local interactions such as close contacts, coordination, and short-range perturbations. The long-range branch is intended to capture broader environmental effects, including solvent-mediated and longer-range electrostatic contributions.

A basic interpolated prediction can be written as

$$\hat{F}_i = \alpha_i \hat{F}_i^{short} + (1 - \alpha_i) \hat{F}_i^{long} \tag{2.3}$$

Where $\alpha_i \in [0,1]$ is the short-scale weight. A larger $\alpha_i$ indicates stronger reliance on the short-scale branch, while a smaller $\alpha_i$ indicates stronger reliance on the long-range branch.

Equivalently, the interpolation can be expressed in residual form:

$$\hat{F}_i = \hat{F}_i^{long} + \alpha_i \left( \hat{F}_i^{short} - \hat{F}_i^{long} \right) \tag{2.4}$$

This interpretation is useful when the long-range branch provides a strong baseline and the short-scale branch acts as a local correction.

## 2.3 Error-supervised hard scale routing

To determine whether the short- or long-range branch is more reliable for each atom, this study defines branch errors:

$$e_i^{short} = \left| \hat{F}_i^{short} - F_i \right| \quad e_i^{long} = \left| \hat{F}_i^{long} - F_i \right| \tag{2.5}$$

A binary route label is defined as

$$y_i^{route} = \mathbf{1}\left[ e_i^{short} < e_i^{long} \right] \tag{2.6}$$

A routing model predicts

$$p_i = P\left( e_i^{short} < e_i^{long} \right) \tag{2.7}$$

The force prediction can be obtained by soft routing,

$$\hat{F}_i = p_i \hat{F}_i^{short} + (1 - p_i) \hat{F}_i^{long} \tag{2.8}$$

and by hard routing,

$$\hat{F}_i = \begin{cases} \hat{F}_i^{short}, & p_i > \tau, \\ \hat{F}_i^{long}, & p_i \leq \tau. \end{cases} \tag{2.9}$$

The route supervision loss is

$$\mathcal{L}_{route} = -\frac{1}{N}\sum_{i}\left[y_i^{route}\log p_i + \left(1 - y_i^{route}\right)\log(1 - p_i)\right] \tag{2.10}$$

This error-supervised routing formulation tests whether branch preference is learnable from available features such as atom type, nearest-ion distance, learned scale weight, and short/long prediction disagreement.

## 2.4 Continuous effective scale interpolation

Hard routing assumes that the optimal prediction lies at one of the two endpoints: $\alpha_i = 0$ or $\alpha_i = 1$.

However, the optimal scale may lie between these endpoints. This study therefore defines a continuous effective scale by searching the interpolation coefficient that minimizes force prediction error:

$$\alpha_i^* = \arg\min_{\alpha\in[0,1]}\left|\alpha\hat{F}_i^{short} + (1 - \alpha)\hat{F}_i^{long} - F_i\right| \tag{2.11}$$

The corresponding continuous oracle prediction is

$$\hat{F}_i^{oracle-cont} = \alpha_i^*\hat{F}_i^{short} + (1 - \alpha_i^*)\hat{F}_i^{long} \tag{2.12}$$

This analysis directly tests whether intermediate effective scales provide additional value beyond hard short/long selection.

A complexity-aware version can be written as

$$\alpha_i^* = \arg\min_{\alpha\in[0,1]}[\mathcal{L}_i(\alpha) + \lambda_C C(\alpha)] \tag{2.13}$$

where $C(\alpha)$ penalizes the cost of using fine-scale information. In the current experiments, this study first sets $\lambda_C = 0$ to focus on prediction performance.

## 2.5 Discrete adaptive scale anchors

To test whether the continuous scale space can be approximated by a compact set of effective scale anchors, this study discretizes $\alpha$ into five levels: $\alpha \in \{0,0.25,0.5,0.75,1.0\}$.

The discrete oracle scale is then defined as

$$\alpha_i^{disc} = \arg\min_{\alpha\in\{0,0.25,0.5,0.75,1.0\}}\left|\alpha\hat{F}_i^{short} + (1 - \alpha)\hat{F}_i^{long} - F_i\right| \tag{2.14}$$

If the discrete oracle approaches the continuous oracle, then a small number of task-induced scale anchors can efficiently approximate the continuous optimal scale space.

This is important because it suggests that adaptive scale discovery does not require exhaustive enumeration of all possible scales. Instead, a small number of learned or loss-selected scale anchors may be sufficient.

## 2.6 Scale pool update and adaptive scale generation

The previous formulation defines effective scales between two existing anchors. This study further extends this into a scale pool update mechanism.

Let the initial scale pool be $\mathcal{S}_0 = \{s_1, s_2, \dots, s_K\}$.
where each scale anchor $s_k$ is associated with a corresponding expert $f_{s_k}$. In the present two-anchor setting, the initial scale pool contains the short-scale and long-range endpoints: $\mathcal{S}_0 = \{s_{short}, s_{long}\}$

Equivalently, when the scale space is represented by the interpolation coordinate $\alpha$, these two endpoints are denoted as $\mathcal{S}_0 = \{0,1\}$, where $\alpha = 1$ corresponds to the short-scale expert and $\alpha = 0$ corresponds to the long-range expert.

For a pair of scale anchors $(\mathrm{s_i}, \mathrm{s_j})$, this study constructs an interpolated expert:

$$f_{ij}^{\alpha}(x) = \alpha f_{s_i}(x) + (1-\alpha) f_{s_j}(x), \quad \alpha \in [0,1] \tag{2.15}$$

The task-effective interpolation coefficient is selected by minimizing the validation or training objective:

$$\alpha_{ij}^{*} = \arg\min_{\alpha \in [0,1]} \left[ \mathcal{L}_{task}\left(f_{ij}^{\alpha}(x), y\right) + \lambda_C C(\alpha) \right] \tag{2.16}$$

The corresponding effective scale is denoted as $s_{ij}^{*} = s_{ij}\left(\alpha_{ij}^{*}\right)$.

If this candidate scale improves the current best objective by more than a predefined threshold, this study inserts it into the scale pool. The improvement is defined as

$$\Delta\mathcal{L} = \mathcal{L}_{best} - \mathcal{L}_{task}\left(f_{ij}^{\alpha_{ij}^{*}}(x), y\right) \tag{2.17}$$

The candidate scale is accepted when $\Delta\mathcal{L} > \epsilon$, and the new effective scale is then inserted into the scale pool:

$$\mathcal{S} \leftarrow \mathcal{S} \cup \left\{s_{ij}^{*}\right\} \tag{2.18}$$

This process can be repeated over neighboring or high-uncertainty scale pairs until no candidate provides sufficient improvement:$\Delta L \leq \epsilon$.

The resulting set $\mathcal{S}^{*} = \{s_1^{*}, s_2^{*}, \dots, s_M^{*}\}$ is interpreted as a task-adapted effective scale set.

The newly generated scale can be used directly as an interpolated expert:

$$f_{s_{ij}^{*}}(x) = \alpha_{ij}^{*} f_{s_i}(x) + \left(1 - \alpha_{ij}^{*}\right) f_{s_j}(x) \tag{2.19}$$

or distilled into a standalone expert:

$$f_{new}(x) \approx \alpha_{ij}^{*} f_{s_i}(x) + \left(1 - \alpha_{ij}^{*}\right) f_{s_j}(x) \tag{2.20}$$

The distillation loss can be written as

$$\mathcal{L}_{distill} = \left| f_{new}(x) - f_{ij}^{\alpha_{ij}^*}(x) \right|^2 \tag{2.21}$$

Thus, scale interpolation becomes scale discovery: the model no longer only combines predefined scales, but can iteratively generate new task-effective scales.

## 2.7 Differentiable scale update

To test whether scale weights can be updated directly by gradient descent, this study treats α as a learnable parameter or a differentiable function of routing features:

$$\alpha_i = g_\phi(z_i) \tag{2.22}$$

where $z_i$ contains atom type, nearest-ion distance, learned scale weight, branch prediction norms, and short/long prediction disagreement features.

The prediction is

$$\hat{F}_i = \alpha_i \hat{F}_i^{short} + (1 - \alpha_i)\hat{F}_i^{long} \tag{2.23}$$

The scale gate parameters $\phi$ are updated by minimizing force loss:

$$\phi \leftarrow \phi - \eta \frac{\partial \mathcal{L}_{force}}{\partial \phi} \tag{2.24}$$

Equivalently, the scale weight itself follows

$$\alpha \leftarrow \alpha - \eta \frac{\partial \mathcal{L}_{force}}{\partial \alpha} \tag{2.25}$$

This study evaluates several differentiable scale update variants:

1. global learnable $\alpha$;
2. atom-type-specific $\alpha$;
3. distance-group-specific $\alpha$;
4. atom-type and distance-group-specific $\alpha$;
5. MLP-based atom-level $\alpha_i$ gate.

This provides a minimal validation that adaptive scale weights can be optimized directly through the task loss.

## 2.8 Extension to Task-Conditioned Scale Pools

The present study focuses on a single molecular force prediction task. However, the proposed adaptive scale refinement framework is not limited to force modeling. In molecular machine learning, different tasks often require different effective modeling resolutions. For example, force prediction may depend on local coordination environments, long-range electrostatics, and solvent-mediated interactions, whereas infrared spectral prediction may depend more strongly on bond-level and functional-group-level vibrational patterns. Docking and binding affinity prediction may emphasize protein-ligand contact regions and binding-pocket environments, while toxicity and ADMET prediction may rely on toxicophore motifs and whole-molecule physicochemical properties.

Therefore, a fixed universal scale pool may be insufficient for general molecular representation learning. Task-conditioned routing is also conceptually related to attention-based and conditional computation mechanisms, where model behavior can depend on input context or task identity [15,16]. Instead, the effective modeling scale should be conditioned not only on the molecular environment, but also on the task objective. This leads to a task-conditioned extension of the current framework.

This study defines a task pool as a set of molecular prediction tasks: $\mathcal{T} = \{t_1, t_2, \dots, t_M\}$

Each task t has its own task-specific loss function: $\mathcal{L}_t^{task}$
The total objective for task t can be written as:

$$\mathcal{L}_t^{total} = \mathcal{L}_t^{task} + \lambda_P \mathcal{L}_t^{physics} + \lambda_C \mathcal{L}_t^{complexity} + \lambda_S \mathcal{L}_t^{scale} \tag{2.26}$$

Here, $\mathcal{L}_t^{task}$ is the supervised prediction loss for task t, $\mathcal{L}_t^{physics}$ represents task-specific physical or chemical constraints, $\mathcal{L}_t^{complexity}$ penalizes unnecessary use of expensive fine-scale experts, and $\mathcal{L}_t^{scale}$ regularizes scale assignment or scale-pool construction.

In the task-conditioned setting, the scale weight is no longer only a function of the local environment. Instead, it also depends on the task identity:

$$\alpha_{i,t} = g_{\phi}(z_i, e_t) \tag{2.27}$$

where $z_i$ denotes local molecular or environmental features for atom, residue, fragment, or sample i, and $e_t$ is a learnable embedding of task t. The two-scale prediction can then be written as:

$$\hat{y}_{i,t} = \alpha_{i,t} f_{short,t}(x_i) + \left(1 - \alpha_{i,t}\right) f_{long,t}(x_i) \tag{2.28}$$

This formulation means that the same molecular environment may use different effective scales under different task objectives.

More generally, for a scale pool containing K candidate scales, the task-conditioned routing probability is defined as:

$$\pi_{i,t,k} = p(s_k \mid x_i, t) \tag{2.29}$$

where $s_k$ is the k-th scale anchor and $\pi_{i,t,k}$ represents the probability of using scale $s_k$ for sample i under task t. The final prediction is obtained by a weighted combination of task-conditioned scale experts:

$$\hat{y}_{i,t} = \sum_{k=1}^{K} \pi_{i,t,k} \, f_{s_k,t}(x_i) \tag{2.30}$$

This formulation generalizes the present short/long interpolation framework into a task-conditioned mixture of scale experts.

The scale pool can also be decomposed into shared and task-specific components:

$$\mathcal{S} = \mathcal{S}_{shared} \cup \mathcal{S}_t \tag{2.31}$$

Here, $\mathcal{S}_{shared}$ *contains general-purpose scales that are useful across multiple tasks, while* $\mathcal{S}_t$ contains task-specific effective scales generated or selected for task t. For example, force prediction may introduce coordination-shell and long-range electrostatic scales, spectral prediction may introduce bond-vibration and functional-group scales, and docking may introduce contact-interface and pocket-environment scales.

A new task-specific scale is introduced only when it provides sufficient loss reduction. If the shared scale pool produces a high task loss : $\mathcal{L}_t(\mathcal{S}_{shared}) > \delta_t$.

Alternatively, a candidate task-specific scale is accepted if it reduces the validation loss of task t by more than a predefined threshold. The task-specific loss reduction is defined as

$$\Delta\mathcal{L}_t = \mathcal{L}_{val,t}(\mathcal{S}_{shared}) - \mathcal{L}_{val,t}(\mathcal{S}_{shared} \cup \{s_{new}^t\}) \tag{2.32}$$

and the candidate scale is inserted when $\Delta\mathcal{L}_t > \epsilon$.
and is then inserted into the task-specific scale pool::

$$\mathcal{S}_t \leftarrow \mathcal{S}_t \cup \{s_{new}^t\} \tag{2.33}$$

This mechanism extends the scale pool update procedure from a single-task setting to a multi-task setting. In the present force-prediction experiments, scale refinement starts from endpoint anchors and generates intermediate effective scales according to force loss. In the generalized framework, different tasks can induce different task-specific scale pools through their own loss functions and constraints.

Thus, scale instability across molecular tasks should not be viewed as a weakness of the framework. Instead, it suggests that the effective modeling scale is a task-induced property. The goal is not to force all molecular tasks to share a single optimal scale, but to learn the conditional distribution of useful scales given both the molecular environment and the task objective: $p(s \mid x, t)$.

This task-conditioned view transforms adaptive scale refinement from a single-task force-prediction method into a more general framework for molecular representation learning.

# 3. Experimental Setup

## 3.1 Dataset and prediction branches

This study uses a NaCl aqueous ionic system as a minimal testbed. Each atom has ground-truth force components and predictions from two fixed branches: short-scale branch and long-range branch.

The goal is not to propose a fully optimized force field, but to test whether adaptive scale selection and scale pool refinement can improve or explain force prediction.

The primary metric is atom-level force MAE averaged over three force components.

### 3.2 Train/test split

For learnable routing and differentiable scale update models, this study uses sample-level group splitting by sample_id. This prevents atoms from the same molecular frame from appearing simultaneously in both training and test sets.

This split is stricter than random atom-level splitting and reduces the risk of overestimating routing performance due to frame-level leakage.

### 3.3 Compared models

This study compares the following models:

1. **Long-only**
   Uses only $\hat{F}^{long}$.
2. **Short-only**
   Uses only $\hat{F}^{short}$.
3. **Fixed-alpha-0.2**
   Uses a globally fixed interpolation coefficient $\alpha = 0.2$.
4. **Learned-alpha-posthoc**
   Uses the learned short-scale weight from the original gate.
5. **Oracle-hard-gate**
   Selects the lower-error branch for each atom.
6. **Oracle-continuous-alpha**
   Searches the best continuous $\alpha \in [0,1]$ for each atom.
7. **Oracle-discrete-alpha-level**
   Restricts $\alpha$ to $\{0, 0.25, 0.5, 0.75, 1.0\}$.
8. **ScalePool-final-update**
   Starts from $\{0,1\}$ and iteratively adds loss-improving intermediate scale anchors.
9. **RFV2-hard / RFV2-soft**
   Error-supervised random forest routing.
10. **HybridV2**
    Uses soft routing in close-contact regimes and hard routing elsewhere.
11. **Gradient-MLP-alpha-gate**
    Uses a differentiable MLP to predict atom-level scale weights and updates them by force loss.

12. **Uncertainty-mid routing**
    Uses an intermediate scale when RFV2 branch preference is uncertain.

# 4. Results

## 4.1 Overall performance

The main overall results are summarized below.

**Table 1. Overall performance of adaptive scale refinement models.**

| Model | Overall MAE | Improvement over Long-only |
|---|---|---|
| Long-only | 399.65 | 0 |
| Fixed-alpha-0.2 | 399.21 | 0.11% |
| Learned-alpha-posthoc | 397.83 | 0.46% |
| RFV2-hard | 396.91 | 0.69% |
| HybridV2-close-RFsoft-else-RFhard | 396.67 | 0.75% |
| Gradient-MLP-alpha-gate | **396.55** | **0.78%** |
| Oracle-hard-gate / ScalePool-{0,1} | 382.67 | 4.25% |
| ScalePool-final-update | 381.23 | 4.61% |
| Oracle-continuous-alpha | **380.96** | **4.68%** |

**Overall atom-level force MAE and relative improvement over the long-range baseline are reported for fixed interpolation, learned routing, differentiable scale gates, oracle routing, and scale-pool refinement models. The best non-oracle result and the best oracle result are highlighted in bold.**

The best non-oracle model is the gradient-updated MLP scale gate. It reduces the overall MAE from 399.65 to 396.55, corresponding to a 0.78% improvement over the long-range baseline.

Although the overall learned improvement is modest, the oracle results reveal a much larger upper bound. This indicates that the short- and long-range branches contain complementary information, but current learnable scale gates only partially recover the oracle potential.

## 4.2 Close-contact regimes show the strongest adaptive-scale benefit

This study stratifies atoms by nearest-ion distance. The most important regime is $d<0.6$ nm, corresponding to close-contact ionic environments.

**Table 2. Performance in the close-contact regime.**

| Model | MAE for d<0.6nm | Improvement over Long-only |
|---|---|---|
| Long-only | 327.22 | 0 |
| Fixed-alpha-0.2 | 314.45 | 3.90% |
| Learned-alpha-posthoc | 303.33 | 7.30% |
| RFV2-hard | 303.31 | 7.31% |
| RFV2-soft | 298.72 | 8.71% |
| Gradient-MLP-alpha-gate | **298.48** | **8.78%** |
| Uncertainty-mid-0.4-0.6 | 298.26 | 8.85% |
| Oracle-hard-gate / ScalePool-{0,1} | 270.45 | 17.35% |
| ScalePool-final-update | 262.04 | 19.92% |
| Oracle-continuous-alpha | **260.51** | **20.39%** |

**Force MAE and relative improvement over the long-range baseline are reported for atoms with nearest-ion distance d<0.6 nm. This regime represents physically difficult local environments where short-range coordination and long-range electrostatic effects may need to be balanced.**

The close-contact regime exhibits much larger gains than the global average. The continuous oracle reduces the close-contact MAE by 20.39%, while the best learnable scale gate recovers approximately 8.78%.

This suggests that adaptive scale refinement is especially useful in physically difficult local regimes where short- and long-range information must be balanced.

### 4.3 Continuous effective scales improve over hard routing

Hard routing chooses either the short-scale or long-range endpoint. However, continuous oracle interpolation achieves lower error than hard routing.

Overall: $Oracle - hard\ MAE = 382.67\ \ Oracle - continuous\ MAE = 380.96$.

In close-contact regimes: $Oracle - hard\ MAE = 270.45,\ \ Oracle - continuous\ MAE = 260.51$.

These results demonstrate that the task-optimal scale is not always one of the predefined endpoints. Instead, intermediate effective scales can provide additional predictive value.

This supports the central hypothesis of this work: fixed human-defined scales should be treated as initial anchors rather than final optimal resolutions.

### 4.4 Discrete adaptive scale anchors nearly approximate the continuous oracle

This study asks whether the continuous scale space can be approximated by a small number of discrete adaptive scale levels: $\alpha \in \{0, 0.25, 0.5, 0.75, 1.0\}$

The five-level discrete oracle achieves an overall MAE of approximately 381.33, close to the continuous oracle MAE of 380.96 and better than the hard oracle MAE of 382.67. In the close-contact regime, the five-level pool reaches approximately 262.56 MAE, close to the continuous oracle value of 260.51.

This indicates that a small number of task-induced effective scale anchors can approximate most of the continuous oracle scale space.

The distribution of oracle discrete scale levels is strongly endpoint-dominated but contains a non-negligible intermediate-scale subset. Most atoms are assigned endpoint scales, but a minority of atoms require intermediate effective scales. These intermediate cases are important because they explain why the discrete oracle improves over the hard oracle.

### 4.5 Loss-guided scale pool update generates intermediate effective scales

To directly validate the proposed scale pool update mechanism, this study initialized the scale pool with two endpoint anchors:

$$\mathcal{S}_0 = \{0,1\}$$

corresponding to long-only and short-only predictions. At each iteration, candidate midpoints between adjacent anchors were evaluated on the training split, and the candidate that provided sufficient loss reduction was inserted into the scale pool.

The scale pool evolved as follows:

**Table 3. Loss-guided scale pool update trajectory.**

| Iteration | Added $\alpha$ | Scale pool | Test MAE |
|---|---|---|---|
| 0 | - | {0,1} | 382.67 |
| 1 | 0.5 | {0,0.5,1} | 381.73 |
| 2 | 0.25 | {0,0.25,0.5,1} | 381.51 |
| 3 | 0.75 | {0,0.25,0.5,0.75,1} | 381.33 |
| 4 | 0.125 | {0,0.125,0.25,0.5,0.75,1} | 381.27 |
| 5 | 0.375 | {0,0.125,0.25,0.375,0.5,0.75,1} | **381.23** |

**The scale pool is initialized with endpoint anchors {0,1}. At each iteration, the candidate intermediate anchor that provides the largest loss reduction is inserted into the scale pool. Test MAE is reported after each update step.**

The final updated scale pool

$$\mathcal{S}_{final} = \{0,0.125,0.25,0.375,0.5,0.75,1\}$$

achieves an overall MAE of 381.23, approaching the continuous oracle MAE of 380.96.

This result shows that adaptive scale refinement is not merely a post-hoc interpolation analysis. Starting from endpoint anchors, loss-guided updates can automatically generate intermediate effective scales and progressively approximate the continuous effective-scale space.

### 4.6 Intermediate scales are sparse globally but enriched in close-contact regimes

The final updated scale pool uses endpoint anchors for most atoms, but intermediate anchors are non-negligible. To compare scale-anchor usage between the global atom set and the close-contact regime, this study visualizes the fraction of atoms assigned to each scale anchor in Figure 2.

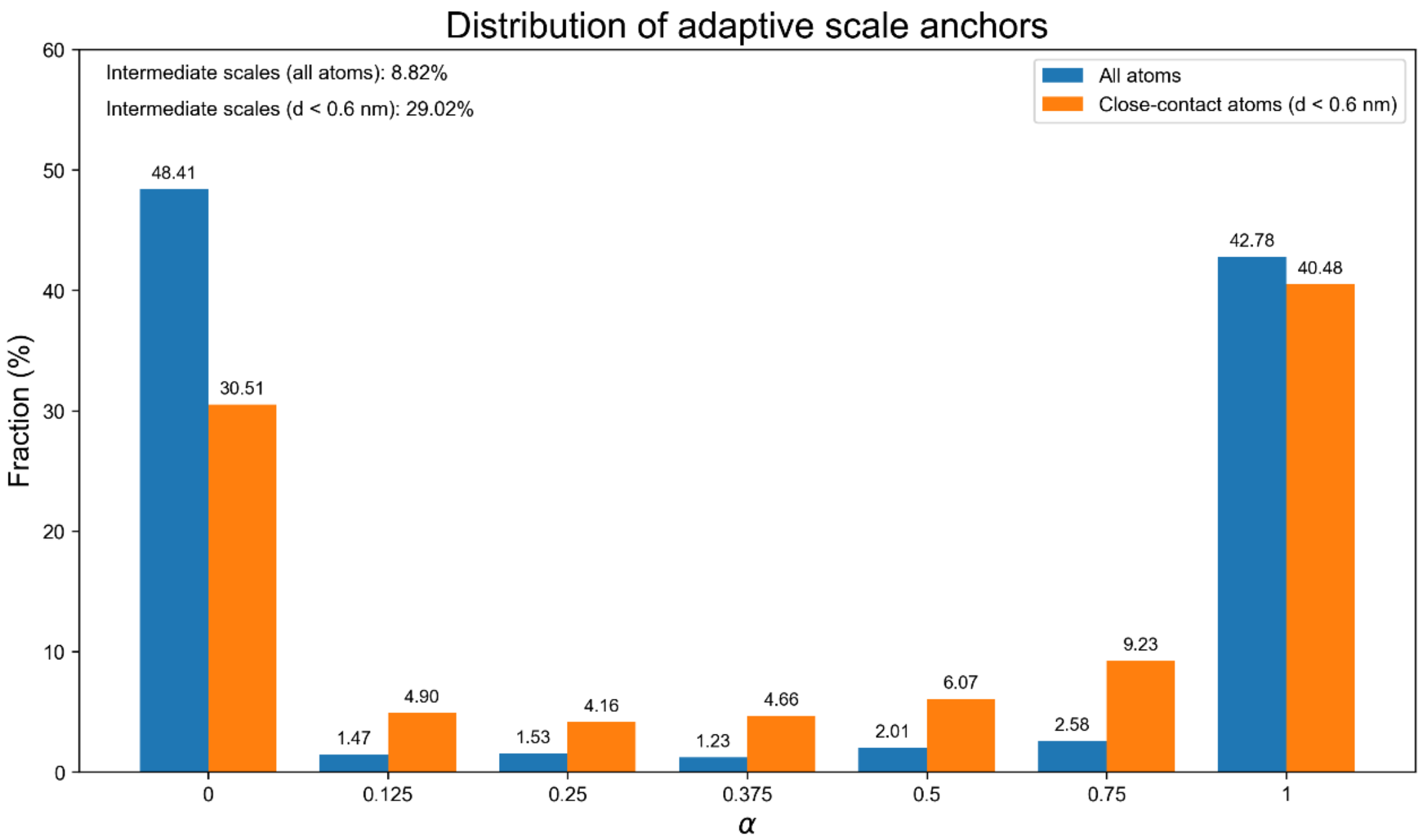

**Figure 2. Distribution of adaptive scale anchors in all atoms and close-contact regimes. Fractions of atoms assigned to each discrete scale anchor α are shown for all atoms and for close-contact atoms with d < 0.6 nm. Endpoint anchors dominate globally, whereas intermediate anchors are substantially enriched in close-contact regimes.**

As shown in Figure 2, endpoint anchors dominate the global distribution. For all atoms, the two endpoint anchors, $\alpha = 0$ and $\alpha = 1$, account for 48.41% and 42.78% of atoms, respectively, while all intermediate anchors collectively account for only 8.82%. In contrast, intermediate-scale usage is substantially higher in the close-contact regime. For atoms with $d < 0.6$ nm, intermediate anchors account for 29.02% of atoms, and the fraction assigned to $\alpha = 0.75$ increases from 2.58% globally to 9.23% in the close-contact subset.

These results indicate that intermediate effective scales are sparse in the global atom population but are strongly enriched in physically difficult close-contact environments. This enrichment supports the interpretation that intermediate scales mainly contribute in transitional local regimes where neither the short-scale nor the long-range endpoint alone is sufficient. It also explains why scale pool updates provide the largest additional gains in close-contact regions.

### 4.7 Loss-guided group-wise scale discovery finds physically meaningful scales

This study evaluated simple loss-guided adaptive scale refinement by searching optimal α values for different groups on the training split and applying them to the test split.

The discovered scale weights are physically interpretable:

**Table 4. Loss-guided group-wise effective scale weights.**

| Group | Effective α |
|---|---|
| global | approximately 0.15 |
| H | approximately 0.15 |
| O | approximately 0.15 |
| Na | approximately 0.65 |
| Cl | approximately 1.00 |
| $d < 0.6$nm | approximately 0.65 |
| $0.6 \leq d < 1.0$nm | approximately 0.00 |
| $1.0 \leq d < 2.0$nm | approximately 0.20 |
| $d > 2.0$nm | approximately 0.05 |

**Optimal group-wise interpolation weights $\alpha$ are searched on the training split and then applied to the test split. The resulting values provide an interpretable estimate of the effective short-scale contribution required by different atom types and distance regimes.**

These results are consistent with chemical intuition. Na and close-contact atoms require larger short-scale contributions, while most water atoms and intermediate-distance regions rely more strongly on long-range context.

This shows that task loss can identify meaningful effective scales without manually assigning them.

## 4.8 Differentiable scale update recovers adaptive scale trends

This study next tested whether the scale weights can be learned by gradient descent instead of post-hoc grid search. This study evaluated global, atom-type, distance-group, atom-distance-group, and MLP-based differentiable α models.

The results show that gradient-based optimization recovers similar trends to grid search. In particular, the distance-group and atom-distance-group scale weights improve over global fixed α, and the MLP gate achieves the best non-oracle performance.

**Table 5. Performance of differentiable scale update variants.**

| Model | Overall MAE | Improvement over Long-only |
|---|---|---|
| Gradient-global-alpha | 399.21 | 0.11% |
| Gradient-atom-alpha | 399.21 | 0.11% |
| Gradient-distance-alpha | 398.37 | 0.32% |
| Gradient-atom-distance-alpha | 398.35 | 0.33% |
| Gradient-MLP-alpha-gate | **396.55** | **0.78%** |

**Overall force MAE and improvement over the long-range baseline are reported for global, atom-type-specific, distance-group-specific, atom-distance-group-specific, and MLP-based differentiable scale update models.**

The learned MLP scale weights are also interpretable:

**Table 6. Mean MLP-predicted scale weights by atom type and distance regime.**

| Group | Mean MLP α |
|---|---|
| All | 0.292 |
| Na | 1.000 |
| Cl | 1.000 |
| H | 0.273 |
| O | 0.327 |
| $d < 0.6$nm | 0.659 |
| $0.6 \leq d < 1.0$nm | 0.072 |
| $1.0 \leq d < 2.0$nm | 0.309 |
| $d > 2.0$nm | 0.276 |

**Mean atom-level scale weights predicted by the MLP scale gate are summarized for atom types and nearest-ion distance groups. These values are used to assess whether the differentiable gate recovers physically meaningful scale preferences.**

The close-contact mean $\alpha \approx 0.659$ closely matches the loss-guided group-wise optimum near 0.65. This provides evidence that differentiable scale update can recover meaningful task-effective scale preferences.

## 4.9 Uncertainty-based intermediate routing provides limited additional benefit

This study also tested a simple uncertainty-based intermediate scale rule. When the RFV2 short-branch probability is low, the model uses long-scale; when it is high, the model uses short-scale; and when it is uncertain, the model uses an intermediate scale $\alpha = 0.5$.

For example:

$$p < 0.4 \Rightarrow \alpha = 0,$$

$$0.4 \leq p \leq 0.6 \Rightarrow \alpha = 0.5,$$

$$p > 0.6 \Rightarrow \alpha = 1.$$

This strategy slightly improves close-contact MAE but does not improve the overall best result. Its best close-contact MAE is approximately 297.67, but its overall MAE remains worse than the Gradient-MLP-alpha-gate.

This suggests that intermediate scales are useful, especially in close-contact regimes, but naive uncertainty thresholds are insufficient for globally optimal scale assignment.

## 5. Discussion

### 5.1 Adaptive scale refinement is different from ordinary multiscale fusion

Ordinary multiscale models usually combine predefined scales, such as atom-level, fragment-level, residue-level, or global representations. In contrast, the present framework treats these predefined scales as initial anchors. The effective scale is then discovered, generated, or updated by task loss.

This distinction is important. The results show that the optimal scale is not always one of the manually chosen endpoints. Continuous oracle interpolation improves over hard routing, discrete adaptive scale anchors nearly approximate the continuous oracle, and scale pool updates automatically generate intermediate anchors from the endpoint pool.

Therefore, adaptive scale refinement should be viewed as a task-induced scale discovery process rather than a fixed multiscale fusion scheme.

### 5.2 Scale pool update transforms scale selection into scale generation

The scale pool update experiment provides a closed-loop validation of the proposed framework. Starting from $\{0,1\}$, the algorithm automatically inserts intermediate anchors, including 0.5, 0.25, 0.75, 0.125, and 0.375. The final pool nearly recovers the continuous oracle performance.

This suggests that scale refinement can proceed without exhaustive enumeration of all possible scales. Instead, the model can begin with a small number of coarse anchors and use loss feedback to generate new effective scales only when they provide meaningful improvement.

In this sense, the framework moves from fixed-scale selection toward adaptive scale generation.

### 5.3 Close-contact regimes are the primary beneficiaries

Across oracle, scale pool update, error-supervised routing, and differentiable scale update experiments, the strongest gains consistently occur in close-contact regimes. This is physically plausible: near ions, local coordination and strong short-range interactions become more important, and a fixed long-range model may be insufficient.

The MLP gate learns high short-scale weights for close-contact atoms, with mean $\alpha \approx 0.659$ for $d < 0.6$ nm. This agrees with the group-wise loss search and supports the interpretation that the learned scale weights reflect meaningful local physical environments.

The anchor usage analysis further shows that intermediate effective scales are enriched in close-contact regimes. Globally, intermediate anchors account for approximately 8.82% of atoms, but in close-contact regimes they account for approximately 29.02% of atoms.

### 5.4 Intermediate scales are sparse but important

The discrete oracle and scale pool update analyses show that most atoms are assigned endpoint scales, but a minority require intermediate scale levels. Although this intermediate subset is relatively small globally, it contributes to the improvement of discrete and continuous oracles over hard routing.

This suggests that intermediate scales may be particularly important for difficult or transitional regimes where neither short nor long branch alone is optimal.

### 5.5 Current learnable gates only partially recover oracle potential

Although the Gradient-MLP-alpha-gate achieves the best performance among non-oracle models, its improvement remains substantially smaller than the continuous oracle upper bound. Specifically, the learned gate improves the overall MAE by 0.78%, whereas the continuous oracle interpolation achieves a 4.68% improvement.

Similarly, in close-contact regimes, the learnable gate recovers approximately 8.78% improvement, compared with 20.39% for the continuous oracle.

This gap indicates that the current feature set and lightweight gate are not sufficient to fully identify the optimal effective scale. Richer local environment descriptors are likely needed, such as:

1.nearest Na distance;

2.nearest Cl distance;

3.coordination numbers within multiple cutoffs;

4.Na-O coordination statistics;

5. Cl-H coordination statistics;

6. local density;

7. RDF shell features;

8. hydration-shell descriptors.

These features may be particularly important for the 0.6–2.0 nm intermediate-distance regimes, where oracle gains remain available but current learned gates show only small improvements.

## 6. Limitations

This study is intended as a minimal validation of adaptive scale refinement rather than a complete force-field model. Several limitations remain.

First, the current experiments use a single NaCl aqueous ionic system. Extension to other salts, concentrations, molecular systems, and biomolecular environments is required.

Second, the short- and long-range branches are treated as fixed prediction sources in most analyses. A fully end-to-end model with jointly trained scale experts and scale gates remains to be developed.

Third, the current MLP gate uses limited routing features. More physically meaningful local structural descriptors may improve scale prediction.

Fourth, complexity regularization is formulated but not fully explored. Future work should test whether adaptive scale selection can reduce computation by invoking expensive fine-scale experts only where needed.

Fifth, the current performance gains are strongest in close-contact regimes, while the overall MAE improvement remains modest. The main value of the current study lies in demonstrating the existence, generation, and learnability of task-effective scales rather than delivering a fully optimized force-prediction model.

Finally, this study should be viewed as an initial validation of the proposed framework rather than a final complete implementation. Future versions will extend the framework to additional ionic systems, molecular environments, jointly trained scale experts, richer local structural descriptors, and task-conditioned scale pools. The accompanying code repository will be used to release implementation details, supplementary scripts, and future experimental updates.

## 7. Conclusion

This study introduced a loss-guided adaptive scale refinement framework for molecular force prediction. The main idea is that molecular representation scales should not be treated as fixed human-defined levels, but as task- and environment-dependent effective modeling resolutions.

Using a NaCl aqueous ionic system, this study showed that short- and long-range force prediction branches have substantial complementarity. Oracle hard routing improves overall MAE by 4.25%, while continuous oracle interpolation further improves it to 4.68%, demonstrating that the optimal scale is not restricted to short/long endpoints. Discrete oracle analysis and scale pool update experiments show that a compact set of adaptive scale anchors can efficiently approximate the continuous oracle scale space. Starting from endpoint anchors {0,1}, the scale pool update mechanism automatically generates intermediate anchors and reaches an overall MAE of 381.23, close to the continuous oracle value of 380.96.

This study further showed that adaptive scale weights can be learned by differentiable optimization. A gradient-updated MLP scale gate achieves the best non-oracle performance, improving overall MAE by 0.78% and close-contact MAE by 8.78%. Although these gains remain far from the continuous oracle upper bound, they demonstrate that scale weights can be updated directly through task loss and can partially translate adaptive scale discovery into predictive improvement.

Overall, these results support adaptive scale refinement as a promising direction for molecular representation learning, force-field modeling, and broader AI-for-science tasks where the optimal modeling resolution is task-dependent.

**Data and Code Availability**

The processed prediction files, scale analysis scripts, oracle scale interpolation scripts, scale pool update scripts, error-supervised routing scripts, and differentiable scale update scripts will be made available upon release of the preprint.

A GitHub repository has been created for code access:

https://github.com/anarantos-lovers/Loss-Guided-Adaptive-Scale-Refinement-for-Molecular-Force-Prediction

The full implementation and supplementary scripts will be uploaded to this repository after the preprint is posted.

**Author Contributions**

L.Y. conceived the study, designed the loss-guided adaptive scale refinement framework, implemented the experiments, analyzed the results, prepared the figures and tables, and wrote the manuscript.

**Acknowledgements**

The author acknowledges the open-source scientific computing community for providing software tools that supported molecular data processing, machine learning experiments, and figure preparation.

## References

[1] Behler, J.; Parrinello, M. Generalized Neural-Network Representation of High-Dimensional Potential-Energy Surfaces. *Physical Review Letters* **2007**, *98*, 146401. https://doi.org/10.1103/PhysRevLett.98.146401.

[2] Smith, J. S.; Isayev, O.; Roitberg, A. E. ANI-1: An Extensible Neural Network Potential with DFT Accuracy at Force Field Computational Cost. Chemical Science 2017, 8, 3192–3203. https://doi.org/10.1039/C6SC05720A

[3] Schütt, K. T.; Sauceda, H. E.; Kindermans, P.-J.; Tkatchenko, A.; Müller, K.-R. SchNet: A Deep Learning Architecture for Molecules and Materials. The Journal of Chemical Physics 2018, 148, 241722. https://doi.org/10.1063/1.5019779.

[4] Unke, O. T.; Meuwly, M. PhysNet: A Neural Network for Predicting Energies, Forces, Dipole Moments, and Partial Charges. Journal of Chemical Theory and Computation 2019, 15, 3678–3693. https://doi.org/10.1021/acs.jctc.9b00181.

[5] Unke, O. T.; Chmiela, S.; Gastegger, M.; Schütt, K. T.; Sauceda, H. E.; Müller, K.-R. SpookyNet: Learning Force Fields with Electronic Degrees of Freedom and Nonlocal Effects. Nature Communications 2021, 12, 7273. https://doi.org/10.1038/s41467-021-27504-0.

[6] Gilmer, J.; Schoenholz, S. S.; Riley, P. F.; Vinyals, O.; Dahl, G. E. Neural Message Passing for Quantum Chemistry. Proceedings of the 34th International Conference on Machine Learning, PMLR 2017, 70, 1263–1272.

[7] Chen, C.; Ye, W.; Zuo, Y.; Zheng, C.; Ong, S. P. Graph Networks as a Universal Machine Learning Framework for Molecules and Crystals. Chemistry of Materials 2019, 31, 3564–3572. https://doi.org/10.1021/acs.chemmater.9b01294.

[8] Gasteiger, J.; Groß, J.; Günnemann, S. Directional Message Passing for Molecular Graphs. International Conference on Learning Representations 2020. arXiv: 2003.03123.

[9] Satorras, V. G.; Hoogeboom, E.; Welling, M. E(n) Equivariant Graph Neural Networks. Proceedings of the 38th International Conference on Machine Learning 2021. arXiv: 2102.09844.

[10] Batzner, S., Musaelian, A., Sun, L. et al. E(3)-equivariant graph neural networks for data-efficient and accurate interatomic potentials. Nat Commun 13, 2453 (2022). https://doi.org/10.1038/s41467-022-29939-5

[11] Batatia, I.; Kovács, D. P.; Simm, G. N. C.; Ortner, C.; Csányi, G. MACE: Higher Order Equivariant Message Passing Neural Networks for Fast and Accurate Force Fields. Advances in Neural Information Processing Systems 2022. arXiv: 2206.07697.

[12] Gasteiger, J.; Giri, S.; Margraf, J. T.; Günnemann, S. Fast and Uncertainty-Aware Directional Message Passing for Non-Equilibrium Molecules. Machine Learning for Molecules Workshop, NeurIPS 2020. arXiv: 2011.14115.

[13] Wu, Z.; Ramsundar, B.; Feinberg, E. N.; Gomes, J.; Geniesse, C.; Pappu, A. S.; Leswing,

K.; Pande, V. MoleculeNet: A Benchmark for Molecular Machine Learning. Chemical Science 2018, 9, 513–530. https://doi.org/10.1039/C7SC02664A.

[14] Yang, K.; Swanson, K.; Jin, W.; Coley, C.; Eiden, P.; Gao, H.; Guzman-Perez, A.; Hopper, T.; Kelley, B.; Mathea, M.; Palmer, A.; Settels, V.; Jaakkola, T.; Jensen, K.; Barzilay, R. Analyzing Learned Molecular Representations for Property Prediction. Journal of Chemical Information and Modeling 2019, 59, 3370–3388. https://doi.org/10.1021/acs.jcim.9b00237.

[15] Shazeer, N.; Mirhoseini, A.; Maziarz, K.; Davis, A.; Le, Q.; Hinton, G.; Dean, J. Outrageously Large Neural Networks: The Sparsely-Gated Mixture-of-Experts Layer. International Conference on Learning Representations 2017. arXiv: 1701.06538.

[16] Vaswani, A.; Shazeer, N.; Parmar, N.; Uszkoreit, J.; Jones, L.; Gomez, A. N.; Kaiser, Ł.; Polosukhin, I. Attention Is All You Need. Advances in Neural Information Processing Systems 2017, 30.